\documentclass[10pt, conference]{IEEEtran}
\IEEEoverridecommandlockouts

\usepackage[english]{babel}
\usepackage{amsmath,amssymb,amsfonts,bm,commath}
\usepackage{algorithmic}
\usepackage{graphicx}
\usepackage{textcomp}
\usepackage{xcolor}

\usepackage[hidelinks]{hyperref}
\usepackage{units}

\newcommand{\mycaption}[2]{\caption[#1]{#1. #2}}

\newcommand{\vect}[1]{\boldsymbol{\bm{#1}}}
\newcommand{\gvect}[1]{\boldsymbol{\mathbf{#1}}}

\newcommand{\vectdot}[1]{\vect{\dot{#1}}}

\newcommand{\mat}[1]{\boldsymbol{\mathbf{#1}}}

\newcommand{\vecthat}[1]{\mathbf{\hat{\text{$\vect{#1}$}}}}

\newcommand{\tcmd}{\mathrm{cmd}}

\newcommand{\vel}{\vect{v}}

\newcommand{\vgait}[1][{}]{\vel_{\mathrm{g}_{#1}}}
\newcommand{\vgaithat}[1][{}]{\widehat{\vel}_{\mathrm{g}_{#1}}}

\newcommand{\pdot}{\vect{\dot{p}}}

\newcommand{\ptd}{\vect{p}_\mathrm{cp}}
\newcommand{\pdtd}{\vectdot{p}_\mathrm{cp}}
\newcommand{\pdtdk}{\vectdot{p}_{\mathrm{cp}_k}}
\newcommand{\pdtdw}{\vectdot{p}_{\mathrm{cp}_w}}
\newcommand{\drive}[1]{{#1}_w}

\newcommand{\mItw}{\mat{I}_{12}}
\newcommand{\mZtw}{\mat{0}_{12}}
\newcommand{\mI}{\mat{I}_3}
\newcommand{\mZ}{\mat{0}_3}
\newcommand{\mi}{\mat{I}_{3}}
\newcommand{\mz}{\mat{0}_{3}}
\newcommand{\noiseQ}[1]{\omega_{#1}}
\newcommand{\noiseR}[1]{\nu_{#1}}
\newcommand{\covGain}{\mat{\xi}}

\newcommand{\Fr}[1][{}]{\vect{f}^{#1}}
\newcommand{\Fri}{\Fr[i]}

\newcommand{\pf}[1][{}]{\vect{p}^{{#1}}}
\newcommand{\pfdot}[1][{}]{\vectdot{p}^{{#1}}}

\newcommand{\pfi}{\pf[i]}

\newcommand{\pfidot}{\pfdot[i]}

\newcommand{\acronym}[1][{}]{{#1}}

\addto\captionsenglish{\renewcommand{\figurename}{Fig.}}
\newcommand{\myfig}[1]{Fig.~\ref{#1}}
\newcommand{\myeq}[1]{\eqref{#1}}

\newcommand{\myeqs}[2]{\eqref{#1} and \eqref{#2}}

\usepackage{eso-pic}

\AtBeginDocument{\AddToShipoutPictureFG*{\AtTextUpperLeft{\put(0,\LenToUnit{8pt}){\parbox{\textwidth}{\centering\sffamily IEEE International Robotic Computing (IRC), Naples, Italy, December 2022.
}}}}}

\begin{document}

\title{State Estimation for Hybrid Locomotion of Driving-Stepping Quadrupeds\\
}

\author{\IEEEauthorblockN{Mojtaba Hosseini}
\IEEEauthorblockA{%
\textit{Autonomous Intelligent Systems} \\
\textit{University of Bonn, Germany}\\
hosseini@ais.uni-bonn.de}
\and

\IEEEauthorblockN{Diego Rodriguez}
\IEEEauthorblockA{\textit{Dexterity Inc.} \\
diego.rodriguez@dexterity.ai}
\and

\IEEEauthorblockN{Sven Behnke}
\IEEEauthorblockA{\textit{Autonomous Intelligent Systems} \\
\textit{University of Bonn, Germany}\\
behnke@cs.uni-bonn.de}
}

\maketitle

\begin{abstract}
Fast and versatile locomotion can be achieved with wheeled quadruped robots that drive quickly on flat terrain, but are also able to overcome challenging terrain by adapting their body pose and by making steps.
In this paper, we present a state estimation approach for four-legged robots with non-steerable wheels that enables hybrid driving-stepping locomotion capabilities.
We formulate a Kalman Filter (\acronym[KF]) for state estimation that integrates driven wheels into the filter equations and estimates the robot state (position and velocity) as well as the contribution of driving with wheels to the above state.
Our estimation approach allows us to use the control framework of the Mini Cheetah quadruped robot with minor modifications.
We tested our approach on this robot that we augmented with actively driven wheels in simulation and in the real world. The experimental results are available at \url{https://www.ais.uni-bonn.de/~hosseini/se-dsq}.
\end{abstract}


\section{Introduction}\label{chapter_introduction}

Robots with legs are well suited for locomotion on irregular terrain.
Walking robots are great for negotiating obstacles like stairs, but they have limited speed and are not very energy-efficient.
In contrast, robots with wheels can move quickly and efficiently on flat terrain and have very low transportation costs---compared to their legged counterparts. They cannot overcome obstacles larger than the radius of their wheels, however.
%
An environment with obstacles or height differences, such as curbs for pedestrians, imposes hard constraints that may be impossible for driving robots to overcome. 

\begin{figure}[tbp]
\centerline{\includegraphics[width=\linewidth]{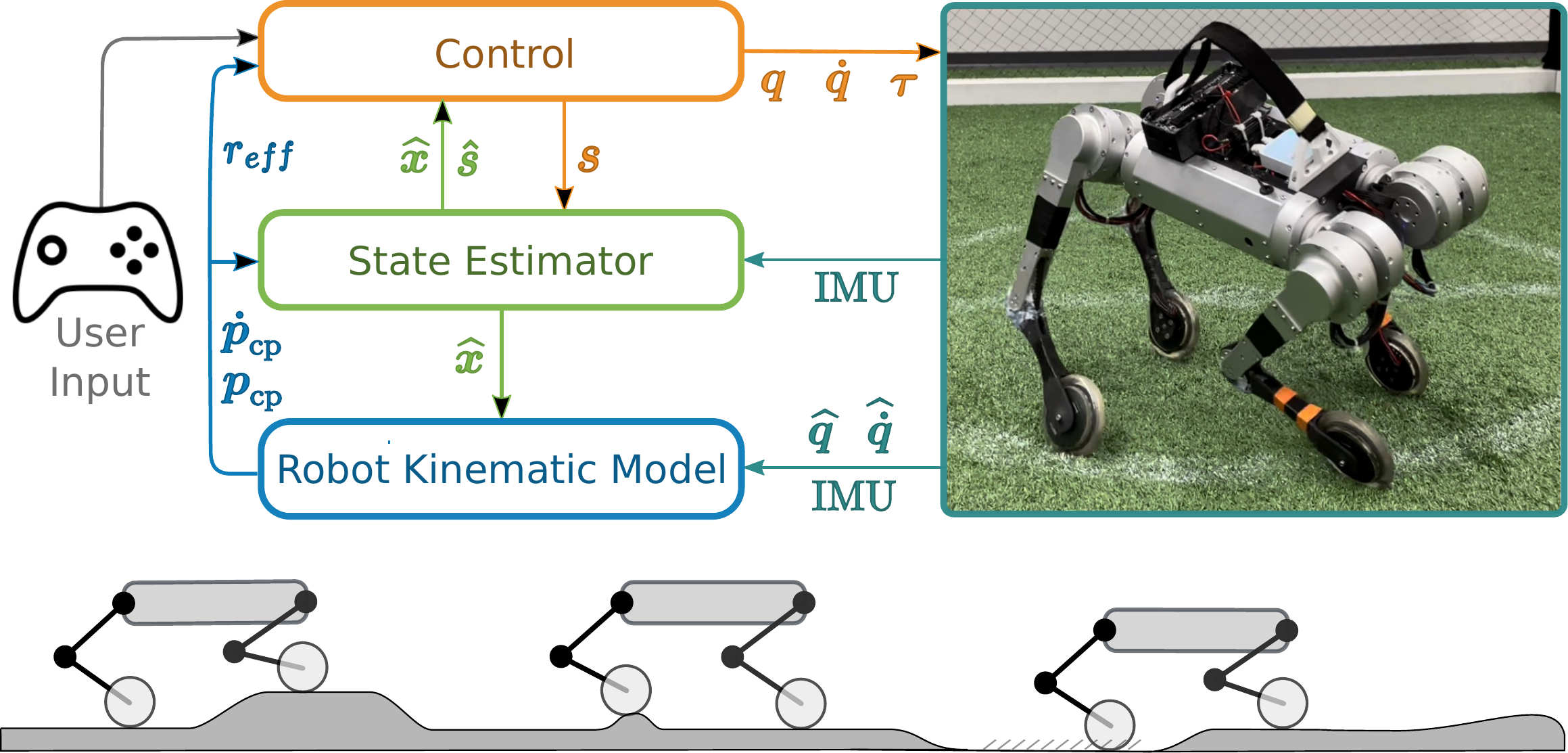}}
\caption{Overview of the control framework (top). The measured joint position $\vecthat{q}$ and velocity $\vecthat{\dot{q}}$,
and data from \acronym[IMU] are sent to the robot model kinematic module, which calculates the position $\vect{p}_\mathrm{cp}$ and velocity $\vect{\dot p}_\mathrm{cp}$ of each contact point as well as the effective radius $r_\mathrm{eff}$ of the end effector. 
The state estimation module uses $\vect{p}_\mathrm{cp}$, $\vect{\dot p}_\mathrm{cp}$, the \acronym[IMU] data and the foot contact states $\vect{s}$ to calculate the robot state $\vecthat{x}$ and the updated contact state $\hat{s}$.
The user input includes the desired driving, stepping, and turning velocities which are used by the control module along with $\vecthat{x}$ to calculate joint position $\vect{q}$, velocity $\vect{\dot q}$, and torque $\gvect{\tau}$.
The wheel controller applies additional control inputs to the wheel joints.
The bottom part illustrates the usefulness of our state estimation for hybrid wheeled-legged quadruped locomotion. The estimated robot height remains unchanged even if the terrain has elevation changes.}
\label{fig_motivation}
\end{figure}

The above characteristics motivated us to combine the capabilities of wheeled and legged robots by using fast and efficient driving on flat terrain and articulated legs to adapt to the terrain and to overcome obstacles by making steps, as illustrated in the bottom part of \myfig{fig_motivation}.
The resulting hybrid locomotion resolves the trade-off between efficiency and speed in systems with wheels and legs by reducing the number of leg swings, since driving is usually preferred over walking on flat terrains. At the same time, stepping is applied to handle steps and irregular terrain. 
This way, we increase the robot's locomotion speed and energy efficiency without compromising its versatility.

In this paper, we present a novel state estimation approach to enable dynamic hybrid driving-stepping locomotion for a wheeled quadruped robot that uses online Model Predictive Control (\acronym[MPC]) and a Whole Body Controller (\acronym[WBC]),
taking into account the additional wheels in terms of their shapes, masses, and inertia tensors.
This approach (\myfig{fig_motivation} top) integrates the wheels into the robot's kinematics model and updates the Kalman Filter's equations.
In summary, this paper makes the following main contributions:
\begin{enumerate} 
\item Integration of wheels into the kinematic model of a hybrid four-legged Mini Cheetah robot to determine the exact contact position and velocity,
\item Incorporation of the corresponding wheel kinematics into the state estimator of hybrid quadrupeds, and
\item Extending the controller to use the state estimator to drive and step simultaneously.
\end{enumerate}

\section{Related Work}\label{chapter_relatedd}
\subsection{Hybrid Locomotion}
Quadruped robots are complex systems that have received much attention in recent years.
Some works also investigated hybrid wheeled-legged, driving-stepping locomotion.
For example, Bjelonic et al.~\cite{anymal_keep_rolling} and Klamt et al.~\cite{ref_13, ref_14} present hybrid driving-stepping methods that use the wheels only for driving while performing the stepping motions separately.
Bellegarda and Byl~\cite{robosimian} and Geilinger et al.~\cite{ref_16}  
 use Trajectory Optimization (\acronym[TO]) to precompute complex trajectories for hybrid locomotion platforms over a time horizon offline. They plan motion on flat terrain by solving a nonlinear programming problem.
De Viragh et al.~\cite{ref_12} extend \cite{anymal_keep_rolling} by computing robot base and wheel trajectories in a single optimization framework using linearized Zero Moment Point (\acronym[ZMP]) constraints.
This approach is only suitable for low update rates, however, and no test on real platforms is performed.
Bjelonic et al.~\cite{anymal_rolling_in_the_deep} present an optimization framework that incorporates the additional degrees of freedom introduced by wheels into the motion generation.
The optimization problem is split into end-effector and base trajectory planning, similar to Rebula et al.~\cite{related_break_down_two1}, to make locomotion planning more manageable for high-dimensional robots with wheels and legs. 
Bjelonic et al.~\cite{paper_3} propose a method that assigns a utility value to each leg to allow the robot to switch between pure driving, static walking, and trotting.
Klemm et al.~\cite{Klemm2019AscentoAT} present the bipedal robot Ascento that drives quickly on wheels, balances on flat terrain, and overcomes obstacles by jumping.
However, the dynamics of the leg links and motors are neglected, which is significant due to the weight of the wheels compared to the body.

In this work, a hybrid driving-stepping Mini Cheetah quadruped robot with actively driven wheels is introduced and used for experiments in simulation and with real hardware.


\subsection{State Estimation}
State estimation plays a crucial role in this type of system.
Robots can localize themselves in their environment using visual or GPS-based methods \cite{kalman_filter_gps, kalman_filter_vision_2, kalman_filter_vision_1}.
However, for a legged robot that lacks perceptual units and maps of the environment, state estimation relies on data from Inertial Measurement Units (\acronym[IMU]s) and the robot's interaction with the environment over multiple intermittent ground contacts.

Lin et al.~\cite{state_est_hexa} fuse
\acronym[IMU] measurements with a leg-based odometer for a hexapod robot.
The state of the robot is calculated based on the assumption that three feet of the hexapod robot are constantly in contact with a flat surface.
Their method is limited to walking and running on three legs and is therefore not applicable to gaits such as trotting where the robot is underactuated.

Blösch et al.~\cite{state_estimation_eth} present an Extended Kalman Filter (\acronym[EKF]) method that fuses kinematic encoder data with \acronym[IMU] measurements by including the absolute position of all feet in the filter state to accurately capture the uncertainties associated with ground contacts.
The resulting filter simultaneously estimates the pose of the main body and the positions of the footholds without making assumptions about the shape of the terrain.
This approach is only tested in simulation, though.

The aforementioned contact-based methods require accurate knowledge of the contact state of each leg and therefore need dedicated sensors on each foot to detect ground contact.
However, the use of accurate contact sensors increases hardware design complexity and associated costs. 
Our proposed approach does not require such sensory capabilities, making it suitable for a broader range of robotic platforms.

Bledt et al.~\cite{state_estimation} present a probabilistic contact estimation algorithm that uses a \acronym[KF] to fuse precomputed probabilities: the contact probability from the gait swing phase, foot height, and foot force.
Similarly, Camurri et al.~\cite{state_estimate_problistic1} estimate the probability of reliable contact and detect foot impacts using internal force sensing.
This information is then used to improve the estimation of the kinematic-inertial state of the robot base, resulting in performance comparable to systems with foot contact sensors.

The state estimation approach proposed in this work is inspired by Katz et al.~\cite{ketz_minicheetah}, who
estimate the position and orientation of a Mini Cheetah quadruped robot for an inertial coordinate system without a perceptual unit.
Instead, information derived from ground contact points is incorporated into the state estimation equations.
In our approach, measurements from an
\acronym[IMU] and joint encoders are fused to account for driving movements. We estimate the 6D robot pose and the position of all four end-effectors using a KF.
The fusion is based on the contact states given by the robot's gait sequence.

\section{Robot Kinematic Model}\label{chapter_kinematic_model}

Accurate kinematic modeling of the robot plays a key role in state estimation.

The Mini Cheetah quadruped has a rubber ball at the end of each leg as an end effector. 
The size and shape of these robber balls are small. Therefore, they have not been previously considered in robot kinematics, but were modeled as single ground contact points~\cite{anymal_rolling_in_the_deep, paper_3, paper_2, anymal_keep_rolling, paper_1}.
In this work, however, the wheels replacing the rubber balls have a radius of 5\,cm, which is non-negligible compared to the shank length of 19\,cm. 
Consequently, we can no longer consider the ground contact points to be in the center of the end effector.

\begin{figure}
	\centerline{
	\includegraphics[width=0.9\linewidth]{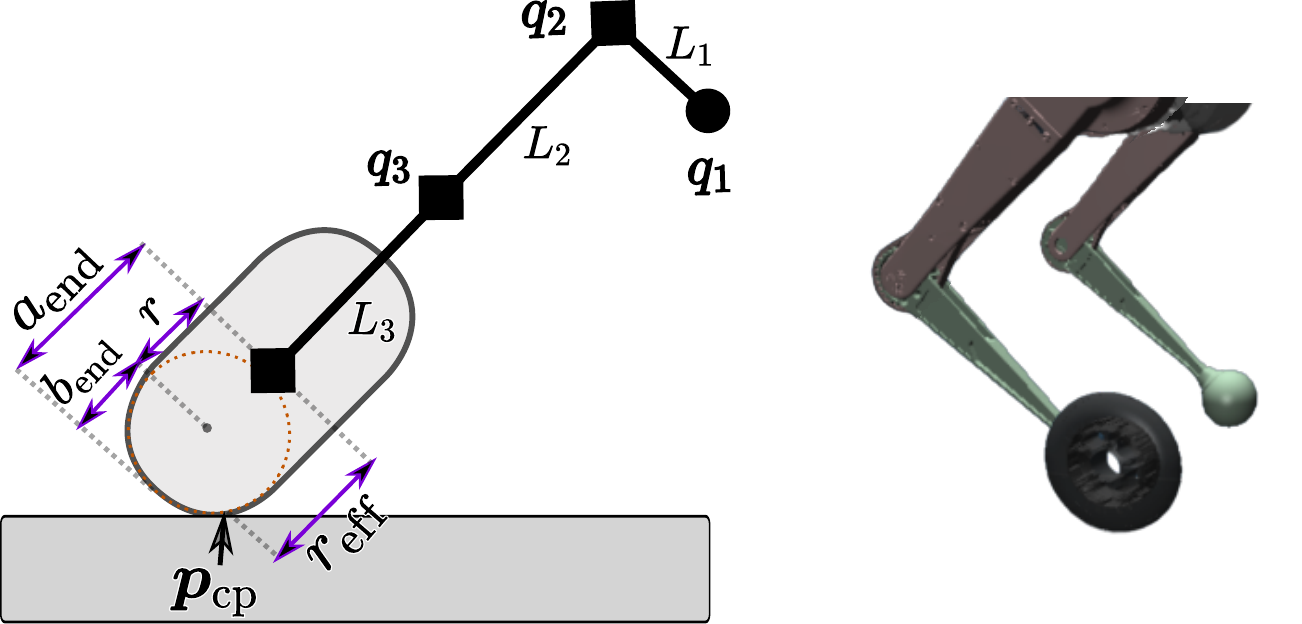}}
	\caption{
 Ground contact point modeling. On the right side, a comparison between the rubber ball and the  wheel end effector is shown. On the left, the joints $q_i$ are shown as circles or rectangles and the links are labeled $L_i$. The radii $a_\mathrm{end}$ and $b_\mathrm{end}$ define the geometry of the wheel end effector, and $\ptd$ is the ground contact position at the effective radius $r_\mathrm{eff}$.}
	\label{kinematics}
\end{figure}

We define a ground contact point as shown in \myfig{kinematics}.
The touchdown position of the end effector on the arbitrary horizontal ground is defined as $\ptd$. 
We assume that the ground surface is locally flat. 
The possible errors between the actual contact point and $\ptd$ are considered negligible in this work.

We model the wheel of the Mini Cheetah robot as shown in \myfig{kinematics}, with radius $a_\mathrm{end}$ and width $b_\mathrm{end}$. 
Assuming that $b_\mathrm{end} \leq a_\mathrm{end}$,
the touchdown position $\ptd$ in the relative coordinate system is defined as:
\begin{equation}
\mbox { \small $
\ptd =
\begin{bmatrix}
	L_3 S_{23} + L_2 S_2    \\
	 L_1 C_1 + L_3 (S_1 C_{23}) + L_2 C_2 S_1  \\
	 L_1 S_1 - L_3 (C_1 C_{23}) - L_2 C_1 C_2   \\
\end{bmatrix} 
-
\underbrace{
\begin{bmatrix}
	0   \\
	 -r S_1 \\
      r C_1 + b_\mathrm{end} \\
\end{bmatrix}}_{\textrm{end effector terms}},
$}
\end{equation}
where $r = a_\mathrm{end} - b_\mathrm{end}$ gives the length of the last link of the kinematic chain, $S_i$ and $C_i$ are abbreviations for $\sin{q_i}$ and $\cos{q_i}$, respectively, $C_{23} = C_2 C_3 - S_2 S_3$ and $S_{23} = S_2 C_3 + C_2 S_3$.
The end effector terms in the equation are the result of an additional end effector geometry in the kinematics that defines the offset of the contact point $\ptd$ from the center of the end effector.

The kinematic contribution of the leg to the touch-down velocity in the world frame is given as:
\begin{equation}\label{p_td_k}
	\pdtdk = \mat{J} \vectdot{q},
\end{equation} 
where 
$\mat{J}$ is the Jacobian matrix of the kinematic chain of the leg including the end effector geometry. The contribution of the added terms for the end effector into the Jacobian matrix $\mat{J}$ is given as:
\begin{equation}
    \mat{J}_w =
\mbox{ \small $ \begin{bmatrix}
	0 & 0 & 0 \\
    r C_1 & 0 & 0 \\
	r S_1 &0 &0
\end{bmatrix}
$}.
\end{equation}

We have addressed the kinematic singularity at joint space by soft-stopping the knee joint before the leg is fully stretched.
We define $r_\mathrm{eff}$ as the effective radius of the end effector (see Eq.~\ref{p_td_k}).
This radius is employed by the state estimator and the wheel controller to accurately convert linear velocities and forces into joint velocities and torques, and vice versa.
Eq.~\ref{p_td_k} is formulated as a function of the hip angle $q_1$ in the world coordinate:
\begin{equation}\label{r_eff}
	r_\mathrm{eff} = a_\mathrm{end} - b_\mathrm{end} \sin{\left(q_1 + \phi\right)},
\end{equation}
where $\phi$ is the roll angle of the robot base.

The contact point velocity $\pdtdk$ given in \myeq{p_td_k} takes into account the robot kinematics and the shape and size of the end effector. 
The robot's pitch velocity $\dot{\theta}$, hip velocity $\dot{q}_2$, and knee velocity $\dot{q}_3$, as well as the robot's roll velocity $\dot{\phi}$ and hip roll velocity $\dot{q}_1$ directly result in additional angular velocity of the end effector (i.e., wheels) along the $x$ and $y$ axes, respectively.
These rotations result in a velocity $\pdtdw$ for the contact point, in addition to the current rotational velocity $\dot{q}_4$ of the wheel read by the encoders. We define $\pdtdw$ as the contribution of the end effector to the global velocity of the contact point:
\begin{equation}\label{p_td_w}
	\pdtdw = \begin{bmatrix} (\dot{\theta} C_1 + \dot{q}_2 + \dot{q}_3 + \dot{q}_4) r_\mathrm{eff} & (\dot{\phi} + \dot{q}_1) b_\mathrm{end} & 0 \end{bmatrix}^\top.
\end{equation}
Finally, the comprehensive contact point velocity $\pdtd$ is:
\begin{equation}\label{p_td}
	\pdtd = \pdtdk + \pdtdw.
\end{equation}
The state estimator makes use of the $\pdtd$, which accurately formulates the velocity at the contact position in world coordinates by taking into account the kinematic shape of the end effector, the velocity feedback from the wheel encoder, and the velocity terms induced by other joints to the end effector.

\section{State Estimation}\label{chapter_state_estimator}

The task of the state estimation module is to estimate the position, velocity, orientation, and angular velocity of the robot base in the world coordinate system. 
Ideally, a state estimator for a quadruped robot should be robust to different types of terrain, gaits, locomotion modes, and travel speeds. It should also take into account the fact that the motions are periodic over an interval and that the system interacts with the environment through multiple intermittent ground contacts.

The control of our four-legged robot uses the residuals between the desired and estimated robot states and therefore relies on accurate robot state estimation. The controller aims to reduce the aforementioned residual and brings the robot state close to the desired one.
However, if the estimated state is incorrect, the controller will generate inappropriate control outputs, resulting in an undesirable body posture that could cause the robot to fall.

The legs in contact with the ground are the reference points for the state estimation. The estimator relies on these reference points and solves for the position and velocity of the body using forward kinematics and Jacobian matrices of the legs in contact.
However, due to obstacles and disturbances, a leg that is supposed to be in contact may actually be in a swinging state, which clearly invalidates the estimation process because the contact condition cannot be trusted.


Our approach for state estimation does not rely on contact or ground reaction force sensors. 
Instead, we use the expected ground contacts defined with a periodic phase-based state planner for different gaits such as Trot, Walk, and Pronk~\cite{state_estimation}.
We change the above-expected contacts to zero when the legs are extended close to their kinematic limits to avoid updating the robot state for legs in singular configurations.

For our state estimation, we use a \acronym[KF] that combines the end effector contact positions \myeq{p_td_k} and the comprehensive contact velocities \myeq{p_td}, the angular velocity of the wheels, and the linear acceleration values from the \acronym[IMU] to obtain a robot position estimate.
Furthermore, we obtain the orientation estimate directly from the \acronym[IMU] hardware.

\subsection{Contact Trust Modeling} \label{section_contact_trust_modeling}

At each instant, each leg of the robot assumes the contact state $s$ with one of two distinct states: contact or swing.
The expected contact state $\hat{s}$ for each leg is determined periodically from the gait at each time step during the control cycle.
In reality, however, the legs are unable to follow the desired commands accurately due to disturbances and surface height differences, which results in a difference between $\hat{s}$ and $s$.
Therefore, simply using the contact locations in state estimation is not applicable and leads to large inaccuracies in the estimated robot state.
For this reason, 
motivated from the phase-based probabilistic model for the expectation of contact given by Bledt et al.~\cite{state_estimation},
we present contact trust modeling that computes a confidence value for the legs expected to be in contact.

\newcommand{\TrustWindow}{\mathcal{W}}

\begin{figure}[tbp]
	\centerline{
	\includegraphics[width=\linewidth]{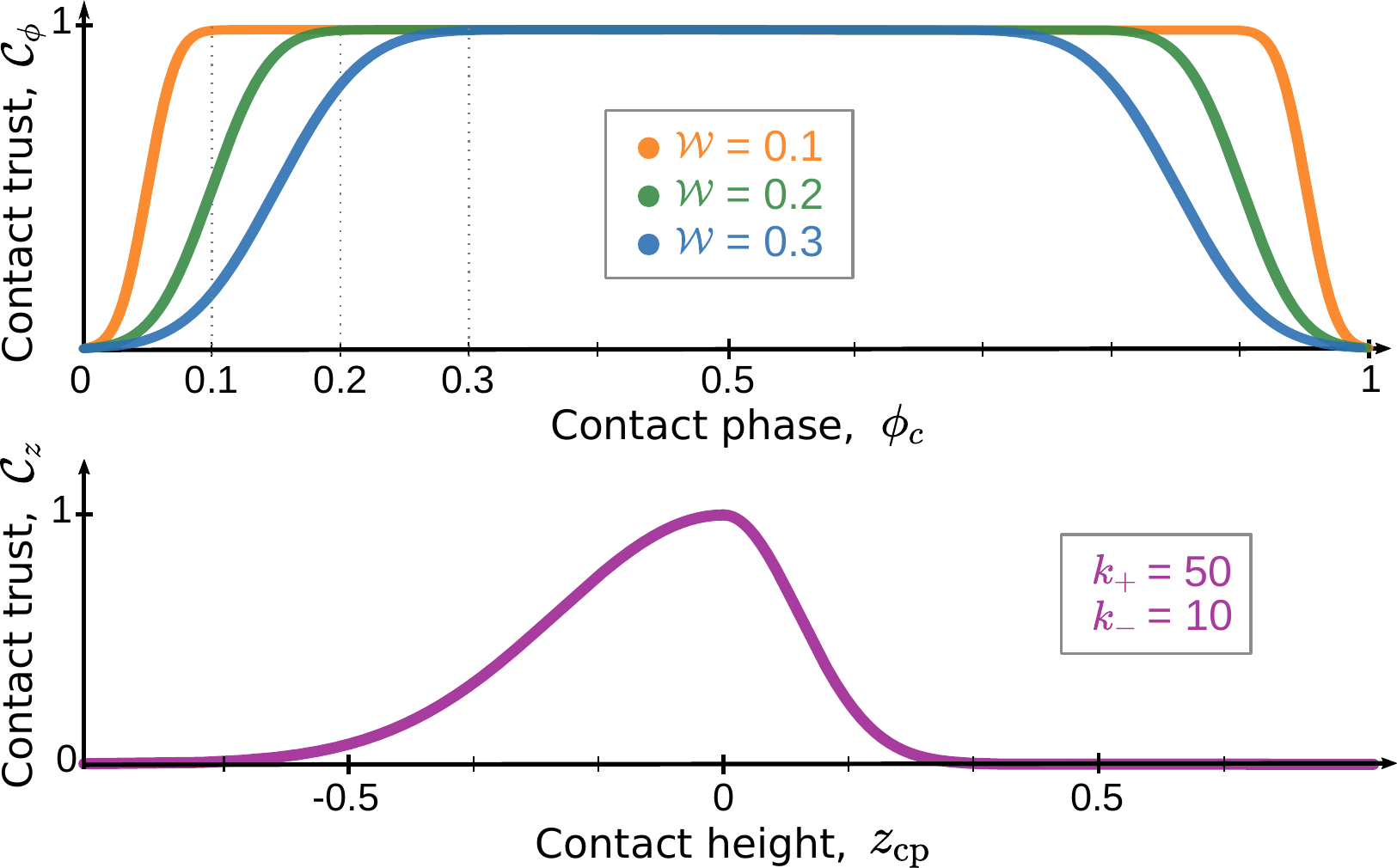}}
	\caption{Contact Trust Modeling.
Top:
The phase-based contact trust $\mathcal{C}_\phi$ is shown for 
various values of $\TrustWindow$ defining the mistrust window at the beginning and the end of the contact phase $\phi_c$.
Bellow:
The contact trust $\mathcal{C}_z$ as a function of contact height $z_\mathrm{cp}$ is presented for distrust gains $k_+$ and $k_-$. Lower distrust gain for negative height differences results in relatively higher trust outputs.
}
	\label{fig_contact_trust}
\end{figure}

The contact phase variable $\phi_{c} \in \left[0, 1\right)$ is defined over the period of the ground contact trajectory as:
\begin{equation} \phi_{c} = \frac{t - t_{c}}{T_{c}},
\end{equation}
which is a linear function of time $t$, where $t_{c}$ and $T_{c}$ are the start time and duration of the contact phase.
Phase-based contact trust (shown in the top part of \myfig{fig_contact_trust}) is formulated as\footnote{Live graph: \url{https://www.desmos.com/calculator/djke61nx8n}}:
\begin{equation}
    \mathcal{C}_\phi=
      \frac{\hat{s}}{2}
      \mbox { $
      \left[\operatorname{erf}\left(4\frac{\phi}{\TrustWindow}-2\right)
      +
      \operatorname{erf}\left(4\frac{1-\phi}{\TrustWindow}-2\right)\right]
      $},\label{eq_contact_trust_phi}
\end{equation}
where $\hat{s}$ chooses the value $1$ and $0$ for the expected contact or swing, respectively, $\TrustWindow$ is the mistrust phase window at the beginning and end of the contact trajectory, and $\mathrm{erf}$ is the sigmoid error function \cite{error_function}.
%

The contact trust as a function of foot height (shown in the bottom part of \myfig{fig_contact_trust}) is defined as:
\begin{equation}
\mathcal{C}_z = \mbox{ \small $
 \begin{cases}
     \exp\left(  -k_+ * {z^2_\mathrm{cp}} \right), & \text{for $z_\mathrm{cp} \ge 0$} \\
     \exp\left(  -k_- * {z^2_\mathrm{cp}} \right), & \text{for $z_\mathrm{cp} < 0$} 
  \end{cases}
  $}\label{eq_contact_trust_z}
\end{equation}
where $z_\mathrm{cp}$ is the contact height and $k_+$ and $k_-$ are the distrust gains for positive and negative contact height values, respectively.
Higher gains lead to lower contact trust for the given foot height. By setting $k_- < k_+$, the lower ground surfaces produce more trustworthy contacts compared to the higher surfaces.


\subsection{State Estimator} \label{section_kf}

To achieve hybrid driving-stepping locomotion, the state estimator should accommodate the wheels that can roll along the $x$-axis.
The \acronym[KF] is able to perform state estimation for robots with legs using the robot measurements and exploiting our contact trust modeling to update its covariance matrices accordingly.

We extended the state estimator of the Mini Cheetah robot~\cite{ketz_minicheetah} to include the wheels in the estimation process accurately.
In the following, we first define the \acronym[KF] state vector and then further explain the \acronym[KF] model and equations.
The notation ${k|k{-}1}$ stands for estimation at time $k$ considering observations up to and including time $k{-}1$.
The identity matrix $\mI$ and the zero matrix $\mZ$ are of size $3\times3$, and $\mat{I}_{12}$ and $\mat{0}_{12}$ are of size $12\times12$.

\subsubsection{Filter State}

The state vector of the filter must be chosen such that the corresponding prediction and measurement equations can be specified consistently.
In our approach, we compose the state vector $\vect{x}$ of the quadruped robot from the position and velocity of the robot base, the contribution of driving to the position and velocity of the robot base, and the contact point positions for all legs relative to the body, which take into account the kinematics of the legs:
\begin{equation}
	\vect{x} := 
	\begin{pmatrix}
		\vect p & \pdot & \drive{\vect p} & \drive{\pdot} & \ptd^{1} \hdots \ptd^{4}\\
	\end{pmatrix},
\end{equation}
where all terms of $\vect x$ are in the world frame, $\vect p$ and $\pdot$ are the global position and velocity of the robot, $\drive{\vect p}$ and $\drive{\pdot}$ are the contribution of driving with the wheels to the position and velocity of the robot, and $\ptd^{i}$ is the ground contact position of the leg $i$ relative to the body (\myfig{kinematics}).

Although the robot dynamics are complex, the above state vector allows us to use a simple linear \acronym[KF].
The inclusion of foot contact positions in the filter state, similar to \cite{state_estimation_eth}, as well as the contribution of driving with wheels into the estimated position and velocity, are key points in our filter design that allow for a simple and consistent representation of the model equations. The \acronym[KF] is able to simultaneously correct for the position of the foot contacts, the pose of the main body, and the contribution of driving. 


\subsubsection{Noise Covariance Update Gain}

When a foot lands on a height difference or at the beginning and end of the contact phase, we should avoid updating the filter for that contact point.
Therefore, we formulate a trust function that takes into account the contact trust values $\mathcal{C}_\phi$ and $\mathcal{C}_z$ from \myeq{eq_contact_trust_phi} and \myeq{eq_contact_trust_z}, respectively, as:
\begin{equation}
\underset{3\times3}{\mathbb{C}} = \mathcal{C}_\phi 
\mbox { \small $
\begin{bmatrix}
    1&0&0\\
    0&1&0\\
    0&0&\mathcal{C}_z\\
\end{bmatrix}
$}.
\label{eq_contact_trust}
\end{equation}
This formulation allows the legs to step on height differences without affecting the estimated robot height.
If the robot steps on a height difference with one or two legs, the estimation avoids considering them in the filter update by assigning a lower trust to the corresponding legs.
However, if multiple legs step on a height difference, the estimator recognizes this height difference as a new ground level and updates the estimated height accordingly.
When the robot steps on negative ground heights, the corresponding legs get stretched and may reach their kinematic limits, which is not ideal. Therefore, for negative height differences, we configure a smaller $k_-$ gain in \myeq{eq_contact_trust_z}, which favors lower ground heights in the estimation, to compute the estimated height of the robot based on the lower surfaces in order to avoid over-extension of the legs.
The trust information in \myeq{eq_contact_trust} is used to define the noise covariance update gain as:
\begin{equation}
    \underset{12\times12}{\covGain} = \mItw + \kappa \left(\mItw - 
        \text{diag} \begin{pmatrix}
		\mathbb{C}^1 & \mathbb{C}^2 & \mathbb{C}^3 & \mathbb{C}^4
	\end{pmatrix}
    \right),
\end{equation}
where $\kappa$ is a hand-tuned value defined as a high suspect number and $\mathbb{C}^i$ is the contact trust value of $i$-th leg in \myeq{eq_contact_trust}.

$\covGain$ encodes the confidence value for all legs (in contact or swing) and is used at each prediction and correction step of the \acronym[KF] to update the process and measurement noise covariance matrices such that only the trustworthy legs contribute to the estimation process.

\subsubsection{Prediction Model}
We formulate prediction equations to propagate the state from one time step to the next. We use the \acronym[IMU] acceleration measurements in the prediction model of the sensor fusion method (similar to \cite{state_estimation_eth, kalman_filter_2}) defined as:
\begin{eqnarray}
	\vect{u} = \mat{R} (\vect{a}_{\mathrm{ IMU }} - \vect{g}),
\end{eqnarray}
where 
$\vect{a}_{\mathrm{ IMU }}$ is the measured linear acceleration in the body frame which is rotated by 
$\mat{R}$ to obtain the acceleration in the global coordinate system.
The standard prediction equations for the \acronym[KF] are as follows:
\begin{align} 
    \vecthat{x}_{k\mid k{-}1} &= \mat{F}_k \vecthat{x}_{k{-}1} + \mat{B}_k \vect{u}_k, \\ 
    \mat{P}_{k\mid k{-}1} &= \mat{F}_k \mat{P}_{k{-}1} \mat{F}_k^\top + \mat{Q}_k,
\end{align}
where $k$ is the time step, $\vecthat{x}$ is the estimate of $\vect{x}$, $\mat{P}$ is the covariance matrix of the state vector, and $\mat{F}$ is the state transition matrix, which applies the system dynamics as:
\begin{equation}
\underset{24\times24}{\mat F} = 
\mbox{ \small $
     \left[
	\begin{array}{cc}
		\begin{matrix}
			\mI & \Delta{t} \mI & \mZ & \mZ  \\
			\mZ & \mI & \mZ & \mZ  			 \\
			\mZ & \mZ & \mi & \Delta{t} \mi  \\
			\mZ & \mZ & \mz & \mi			 \\
		\end{matrix} & \hspace{.5em} \mZtw \hspace{.5em}  \vspace{.25em}\\
		\mZtw &  \mItw \vspace{.25em}
	\end{array}
    \right] $}.
\end{equation}
The matrix 
$
\underset{24\times3}{\mat{B}} =
    \begin{bmatrix}
        \mZ & \Delta{t} \mI & \mZ & \hdots & \mZ 
    \end{bmatrix}^\top 
 $
is defined to update the velocity of the body from the acceleration vector $\vect u$,
and $\mat{Q}$ is the covariance matrix of the process noise
defined as:
\begin{equation}
  \mbox{ \small $
    \underset{24\times24}{\mat Q} = 
    \begin{bmatrix}
        \begin{matrix}
            \noiseQ{\vect{p}} \mI & \mZ & \mZ &\mZ \\
            \mZ & \noiseQ{\pdot} \mI & \mZ &\mZ \\
            \mZ & \mZ & \noiseQ{\drive{\vect{p}}} \mI &\mZ \\
            \mZ &\mZ & \mZ & \noiseQ{\drive{\pdot}} \mI \\
        \end{matrix} & \hspace{1.5em} \mZtw \hspace{1.5em} \vspace{0.75em}\\
        \mZtw & \noiseQ{\ptd} \hspace{0.1em} \covGain \hspace{0.1em} \mItw \vspace{0.75em}
    \end{bmatrix}
    $},
\end{equation}
where the $\noiseQ{}$ parameters are the process noise values for the corresponding variables in the state vector $\vect{x}$.
The white noise terms $\noiseQ{\ptd}$ account for some degree of slippage of the feet, and $\covGain$ sets the noise parameter of the swinging feet to a very large value, allowing the corresponding foot to shift and reset its position estimate when it makes contact with the ground again.
This process deals with the contact changes required to execute a gait.

\subsubsection{Correction Model}

Knowing that the prediction in the hybrid locomotion scheme is likely to contain inaccuracies, we can use available measurements to correct the prediction and obtain a more educated estimate for the position. Since the wheels are integrated into the hybrid quadrupedal robot, the measurement vector $\vect{z}$ contains the contact position and kinematic contribution, as well as the driving contribution to the contact point velocity for all legs as follows:
\begin{equation}
	\vect{z} := \left(
	\begin{array}{ccc}
		\ptd^{1} \hdots \ptd^{4} & \pdtdk^{1} \hdots \pdtdk^{4} &\pdtdw^{1} \hdots \pdtdw^{4}
	\end{array}\right),
\end{equation}
where all terms of $\vect{z}$ are in the global frame, $\ptd^{i}$ is the measured contact position of leg $i$ relative to the body, $\pdtdk^{i}$ and $\pdtdw^{i}$ are the measured linear contact point velocity from kinematics and driving from Eq. \myeqs{p_td_k}{p_td_w}, respectively.

The separate inclusion of the kinematic and driving contributions in the measurement vector $\vect{z}$ plays a key role in the estimation to correctly capture the effects of stepping and driving on the position of the hybrid quadruped robot.
The correction equations of the standard \acronym[KF] are formulated as follows:
\begin{eqnarray}
	&\vect{\widetilde{y}}_k &= \vect{z}_k - \mat{H}_k \vecthat{x}_{k\mid k{-}1},\\
	&\mat{S}_k &= \mat{H}_k \mat{P}_{k \mid k{-}1} \mat{H}_k^\top + \mat{R}_k,\\
	&\mat{K}_k &= \mat{P}_{k\mid k{-}1} \mat{H}_k^\top \mat{S}_k^{{-}1},\\
	&\vecthat{x}_{k\mid k} &=\vecthat{x}_{k\mid k{-}1} + \mat{K}_k\vect{\widetilde{y}}_k,\\
	&\mat{P}_{k\mid k} &= (\mat{I} - \mat{K}_k \mat{H}_k) \mat{P}_{k\mid k{-}1},
\end{eqnarray}
where $\vect{\widetilde{y}}$ is the measurement residual (or innovation), $\mat{S}$ is the covariance of the residual (or innovation covariance), and $\mat{K}$ is the Kalman gain.

$\mat{R}$ is the measurement noise covariance matrix that encodes the confidence we place in the accuracy of the measurements. It is defined as follows:
\begin{equation}
	\underset{36\times36}{\mat{R}} =\covGain
	\begin{bmatrix}
		\noiseR{\ptd} \mItw & \mZtw & \mZtw \vspace{0.4em} \\
		\mZtw & \noiseR{\pdtdk} \mItw & \mZtw \vspace{0.4em} \\
		\mZtw & \mZtw & \noiseR{\pdtdw} \mItw
	\end{bmatrix} ,
\end{equation}
where the $\noiseR{}$ parameters are the measurement noise values for the corresponding variables in $\vect{z}$.
The noise covariance update gain $\covGain$ is applied to the measurement confidence. Similar to the process noise covariance, $\covGain$ sets the noise parameter of the untrusted feet to a very large value when they are not in contact with the ground, which prevents the corresponding measurements from participating in the estimation process. Hence, we can rotate the wheels at different speeds during the swing phase without affecting the estimation process.

We construct the observation matrix $\mat{H}$ with the following Jacobian:
\newcommand{\Mt}[1]{\mbox{ $ #1 $ }}
\newcommand{\Mtt}[1]{\mbox { \footnotesize $ \begin{array}{cccc} #1 \end{array} $} }
\begin{equation}
 \underset{36\times24}{\mat H} = \left[
  \mbox{ \small $
    \begin{array}{cc}
        \vspace{0.5em} \Mtt{
            \mI & \mZ & -\mi & \mz \vspace{-0.25em}\\
            \vdots & \vdots & \vdots & \vdots} 
            & \hspace{-0.5em}  \vspace{0.05em} -\mat{I}_{12} \hspace{0.5em}  
            \\
        \vspace{0.5em} \Mtt{
            \mZ & \mI & \mz & -\mi \vspace{-0.25em}\\
            \vdots & \vdots & \vdots & \vdots} 
            & \hspace{-0.75em} \vspace{0.05em} \mat{0}_{12} 
            \\
        \Mtt{
            \mZ&\mZ&\mz& \hspace{0.65em} \mi \vspace{-0.25em} \\
            \vdots & \vdots & \vdots & \vdots} & \hspace{-0.75em} \mat{0}_{12}
    \end{array}
  $}
\right].
\end{equation}

\section{Control}\label{chapter_control}

The control method of our quadruped robot is based on Model Predictive and Whole-Body Controllers (MPC and WBC), which rely on accurate state estimation of the robot.
Both \acronym[MPC] and \acronym[WBC] directly use the residual between the desired and estimated robot positions, and aim to reduce this residual and bring the robot close to the desired state.
The \acronym[MPC] generates reaction force commands that are enhanced with the \acronym[WBC] to achieve accurate and fast joint control and refined reaction forces \cite{paper_2}.

\subsection{Gait Controller}

Our state estimator estimates the robot's position and velocity and separately includes the contribution of driving with wheels to the robot's estimated position and velocity. This separation allows us to use the Mini Cheetah control scheme with minor modifications and execute arbitrary gaits while simultaneously driving with wheels.
By taking the current velocity of the robot without the contribution of driving as:
\begin{equation}
\vgaithat = \vectdot{p} - \drive{\pdot},
\label{eq_v_g}
\end{equation}
we update the footstep planner equations in \cite{paper_2}:
\begin{align}
	&\vect{p}_\mathrm{target} = \vect{p}_\mathrm{symmetry} + \vect{p}_\mathrm{centrifugal}, \label{eq_p_cmd} \\
	&\vect{p}_\mathrm{symmetry} =\frac{t_\mathrm{stance}}{2} \vgaithat
	+ g_v \left( \vgaithat - \vgait^\tcmd \right), \label{eq_p_symmetry} \\
	&\vect{p}_\mathrm{centrifugal} = \frac{h}{2g} \vectdot{p} \times \gvect{\omega}, \label{eq_p_centrifugal}
\end{align}
where $t_\mathrm{stance}$ is the duration of the contact phase, $\vectdot{p}$ and $\gvect{\omega}$ are the linear and angular velocities of the robot at time step $k$, $g_v$ is the velocity feedback gain, and $\vgait^\tcmd$ is the stepping velocity command.

Eq. \myeq{eq_p_symmetry} allows the controller to execute a gait without the influence of driving with wheels, and \myeq{eq_p_centrifugal} applies a centrifugal foot offset that accounts for the total robot velocity (driving and walking together), resulting in a well-balanced robot when driving with wheels and simultaneously turning. 

Since the wheels are symmetric, we assume constant inertia for the rotating wheels and update the dynamic model of the robot by including the mass and inertia of the wheels in the lower leg linkage and relying on the WBC to correctly compute the joint control commands.

\subsection{Wheel Controller}

While the robot is driving, the feet in the contact phase may not follow the desired rolling velocity due to interference or inadequate control along the rolling direction of the wheels, resulting in foot errors.
We add a corrective velocity along the rolling direction to each foot during the contact phase, which reduces the foot position error and thus increases the control performance. The corrective velocity is defined for the $i$-th foot as follows:
\begin{equation} \label{eq_kinematic_leg_correction}
	\pfidot_\mathrm{corrective} =
	\begin{bmatrix}
		k_p &0&0
	\end{bmatrix}^\top 
	\odot
	\pfi_\mathrm{error},
\end{equation}
where
$k_p$ is the configured p-gain for the rolling direction, $\odot$ is the element-wise multiplication operator, and $\pfi_\mathrm{error}$ is the filtered position error between the current and desired foot positions.

The wheel joints are controlled by a DI controller that follows the input velocities. However, the low resolution of the encoders (84 pulses per revolution) does not allow perfect velocity control, resulting in unwanted wheel motion and reducing the effectiveness of the controller.
We solve this problem by converting the reaction forces from the \acronym[WBC] to the wheel joint torque $\tau$ and passing them as feed-forward torque commands to the joint controller, which makes the wheels work against the reaction forces.
The $i$-th wheel angular velocity $\dot q$ and torque $\tau$ are given by:
\begin{align}
	&\tau =\begin{bmatrix}
			r^i_\mathrm{eff}&0&0
		\end{bmatrix}^\top \cdot \mat{R} \Fri , \label{eq_wheel_tau} \\
	&\dot{q} =\begin{bmatrix}
			\frac{1}{r^i_\mathrm{eff}}&0&0
		\end{bmatrix}^\top \cdot \mat{R} \left( \pfidot_\tcmd + \pfidot_\mathrm{corrective} \right), \label{eq_wheel_q_dot}
\end{align}
where 
$r_\mathrm{eff}$ is the effective radius of the wheel in \myeq{r_eff},
$\mat{R}$ is a rotation matrix for conversion from global to the body reference frame,
$\Fri$ is the current reaction force from \acronym[WBC], and $\pfidot_\tcmd$ is the commanded foot velocity from the footstep planner \cite{paper_2}.

Since the wheel joint velocities are set according to the foot velocities in \myeq{eq_wheel_q_dot}, the tracking of swings along the rolling direction is greatly improved.
If the wheel collides with a surface during the swing, it will continue to turn and help the foot bypass small height differences without additional swings.

\section{Experimental Results}\label{chapter_evaluation}
We extended the Mini Cheetah's hardware to a hybrid quadruped by designing and fabricating relatively small wheels and enhanced shank links.
These wheels can generate a maximum torque of \unit[2.1]{Nm} and a speed of \unit[2150]{rpm}, theoretically allowing the robot to reach up to \unitfrac[40]{km}{h}.
We use ODrive\footnote{\url{https://odriverobotics.com}} BLDC drivers with custom firmware to control the speed and torque of the wheels with a low-resolution encoder (Hall sensor with 84 pulses per revolution).
The modifications increase the weight of each leg by \unit[0.39]{kg}, bringing the total weight of the extended robot to \unit[12.5]{kg} (from its original weight of \unit [10]{kg}).
We transferred the open-source software version of the above robot to the ROS ecosystem and employed MuJoCo as a multibody simulator \cite{todorov2012mujoco} that can accurately simulate the contact dynamics and rolling friction of the additional wheels, as well as the joints and corresponding rotor dynamics.

We intend to use the resulting system to study hybrid locomotion and associated control strategies. It will also serve as a research platform for other desired areas of hybrid mobile robotics, including navigation, path planning in complex terrain, and the application of learning-based control algorithms.
The video showing the evaluation and performance of the real and simulated robot is publicly available\footnote{\url{https://www.ais.uni-bonn.de/~hosseini/se-dsq}}.

\begin{figure}[tbp]
	\centerline{
	\includegraphics[width=\linewidth]{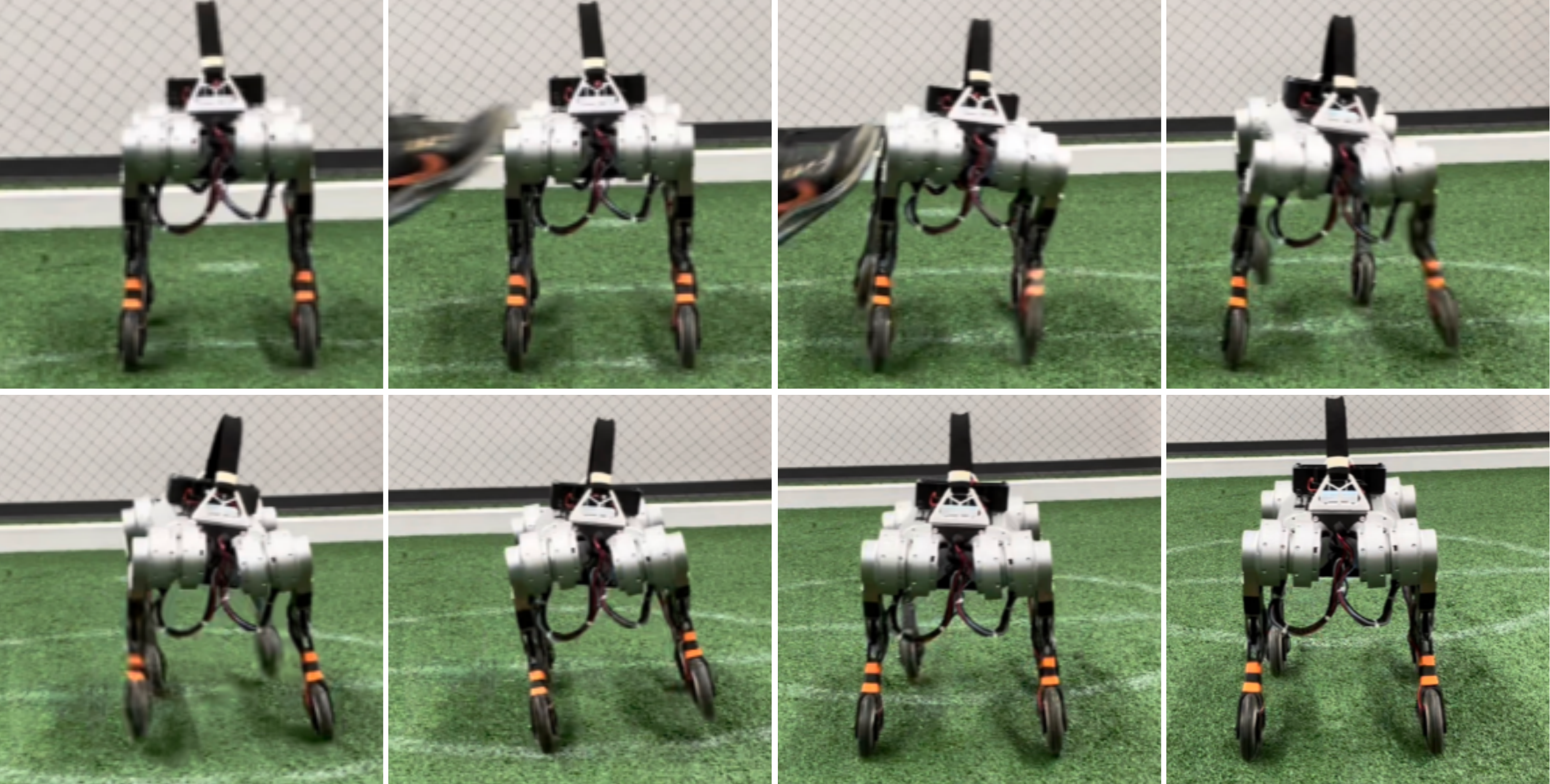}}
	\caption{Hybrid driving-stepping performance.
This figure shows the robot performing forward hybrid driving-stepping with a trot gait (the robot is held in the center of the image and the background shifts backward). When the robot is disturbed, it performs side steps to balance the robot and maintain its current driving state.
	}
	\label{eval_drive_assistant_disturb}
\end{figure}

Hybrid driving is achieved by the robot exhibiting a natural driving style and stepping. The controller steers the robot along the rolling direction of the wheels and minimizes the length of swings along the $x$ axis in the local frame.
Our modified robot achieved a velocity of \unitfrac[20]{km}{h} while driving with wheels.
\myfig{eval_drive_assistant_disturb} demonstrates the hybrid driving-stepping, which allows the robot's agile stepping to take control of the perturbed robot, maintain its balance, and perform corrective steps to follow the commanded track as the robot continues driving.

\begin{figure}[tbp]
	\centerline{
	\includegraphics[width=\linewidth]{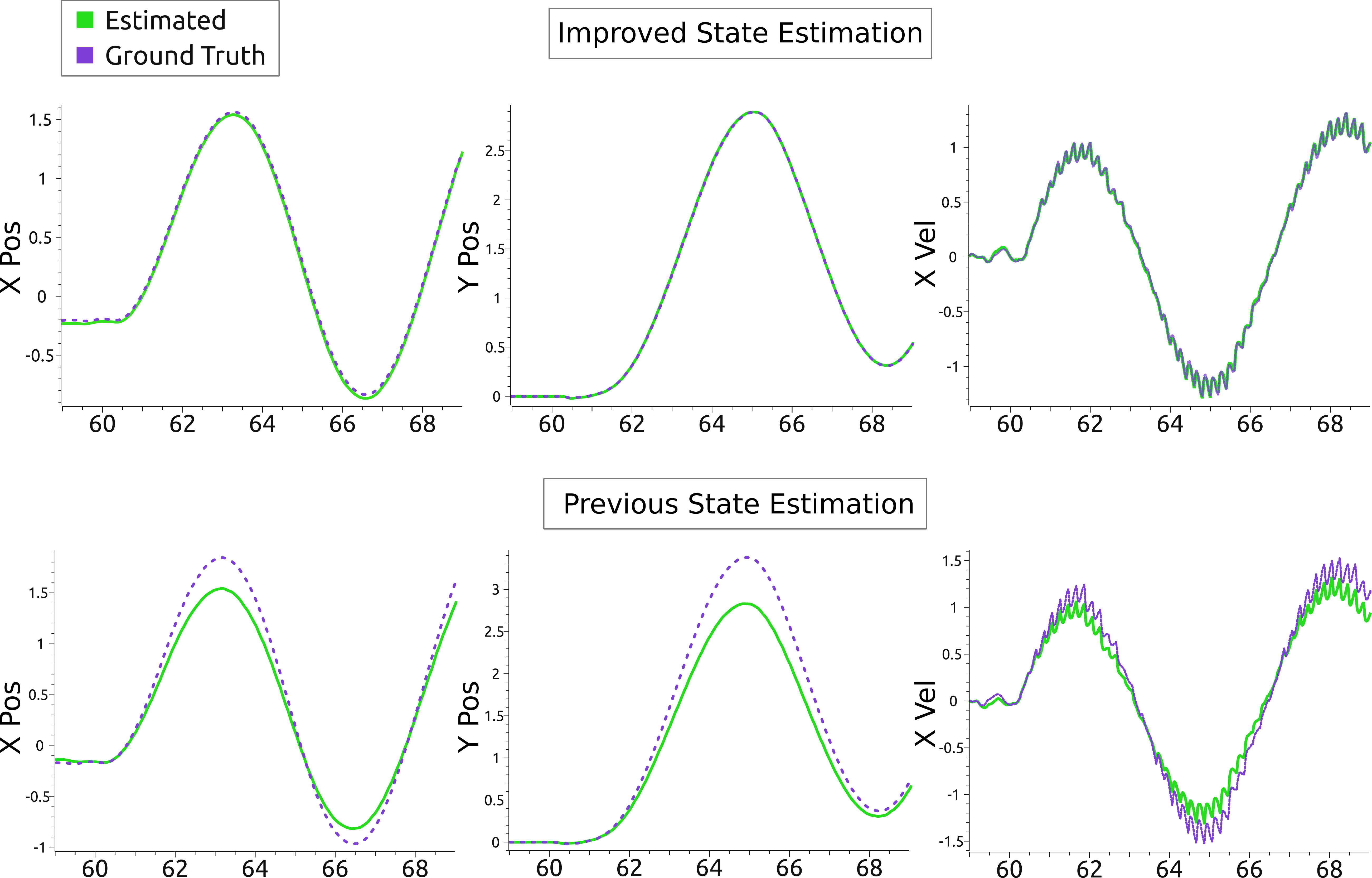}}
	\mycaption{State estimator evaluation graph}{
		We compare the accuracy of state estimation in tracking with the true values (in the simulator) for our method enabled in the top row and the published version~\cite{ketz_minicheetah} for the Mini Cheetah with our method disabled in the bottom row. In this experiment, the robot walks forward and turns left at the same time.
The shape of the end effector needs to be modeled, but in previous work, the center of the rubber ball was assumed to be the contact point and its shape was ignored.
		With our method, the exact ground contact point between the end effector and the ground is calculated, and the effect of the angular velocity of the body and the velocity of other joints on the velocity of the ground contact point is accurately calculated to achieve accurate position and velocity tracking for our robot.
	}
	\label{eval_state_estimate_wheel}
\end{figure}

The tracking accuracy of the state estimator is demonstrated in \myfig{eval_state_estimate_wheel}, which compares our contribution to the state estimator with previous work by Katz et al.~\cite{ketz_minicheetah}.
We set the robot in the simulator to walk forward and turn left at the same time, and compared the true value received from the simulator with the estimated values for the robot's position and velocity.
The previous method ignores the shape of the end effector assuming that its size is negligible. However, the rubber ball at the end of each leg has a radius of \unit[2.5]{cm}, which is considerable compared to the \unit[19]{cm} length of the shank link.
The errors in the position and velocity of the legs resulting from the above assumption accumulate over time and reduce the accuracy of the estimate.
The above errors become even more pronounced for wheels with a radius of \unit[5]{cm}. Our method accurately formulates the position and velocity of the ground contact point, taking into account the shape of the wheel and the effect of other joint velocities, and the angular velocity of the robot on the velocity of the contact point, resulting in an accurate position and velocity estimates, as shown in \myfig{eval_state_estimate_wheel}.

\begin{figure}[tbp]
	\centerline{
	\includegraphics[width=\linewidth]{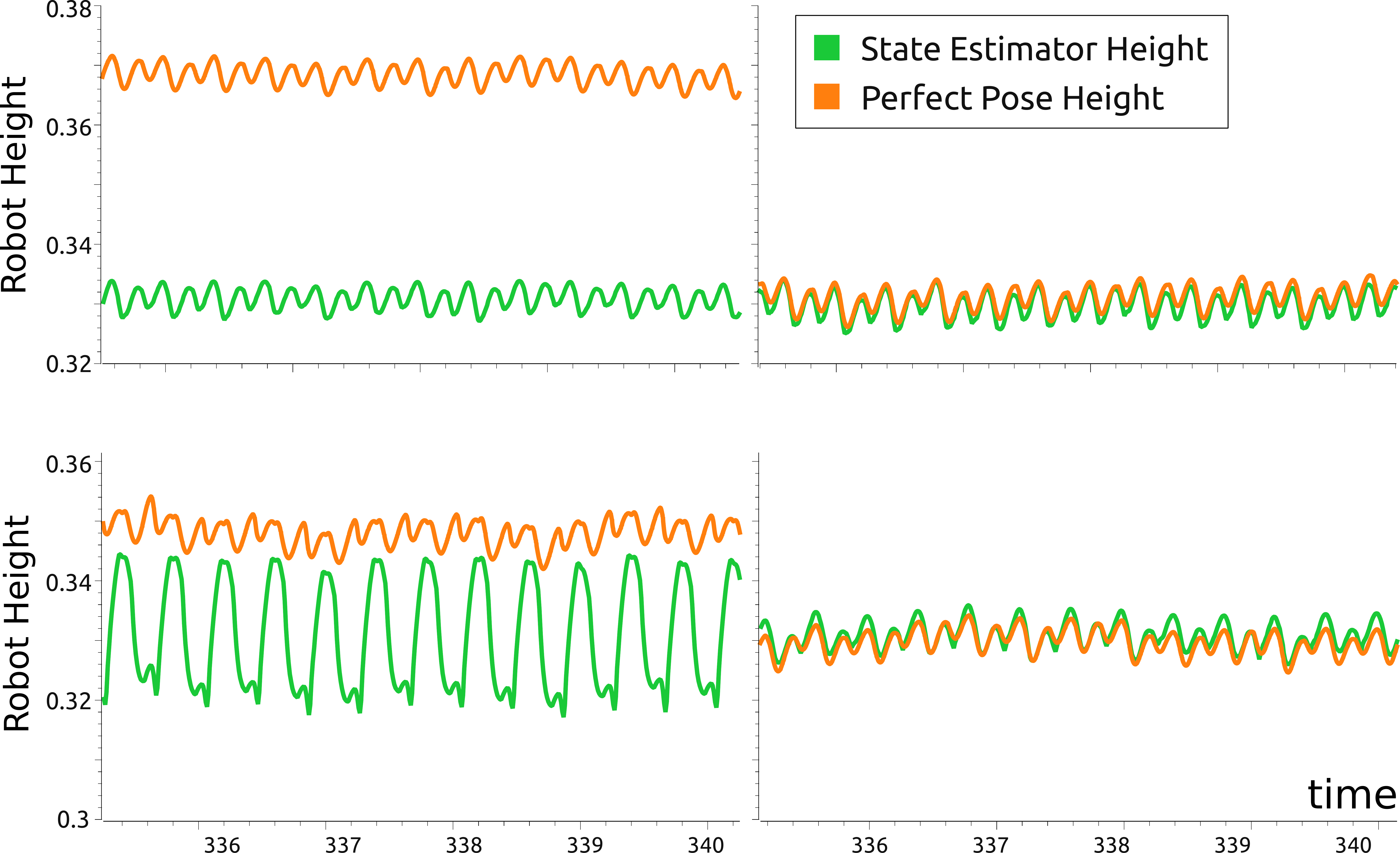}}
	\caption{Estimated height comparison. The green and orange graphs show the estimated and actual heights of the robot, respectively. The graphs in the top row refer to the situation where the robot steps with both front feet on an obstacle of height \unit[0.08]{m}.
 The graphs in the left column show the state estimator output without applying the distrust gain formulation given in \myeq{eq_contact_trust}, and in the right column with applying the distrust gain. With our approach, the height of the robot remains unchanged even when stepping on the height difference.
 	}
	\label{eval_ss}
\end{figure}

We evaluated the effect of the trust formulation in Eq. \myeq{eq_contact_trust} in a scenario where the robot places one or two legs on an obstacle with a height of \unit[8]{cm}. Our approach successfully captures the difference and applies the distrust gains to modify the KF noise and process covariance matrices, resulting in a near-perfect estimate of the robot height. However, previous work has failed to find a suitable corresponding 
 estimation. \myfig{eval_ss} shows this evaluation. The high oscillation range in the lower-left graph for the estimated height is the result of the trotting gait of the robot, where the front right leg periodically steps on the obstacle.

\begin{figure}[tbp]
	\centerline{
	\includegraphics[width=\linewidth]{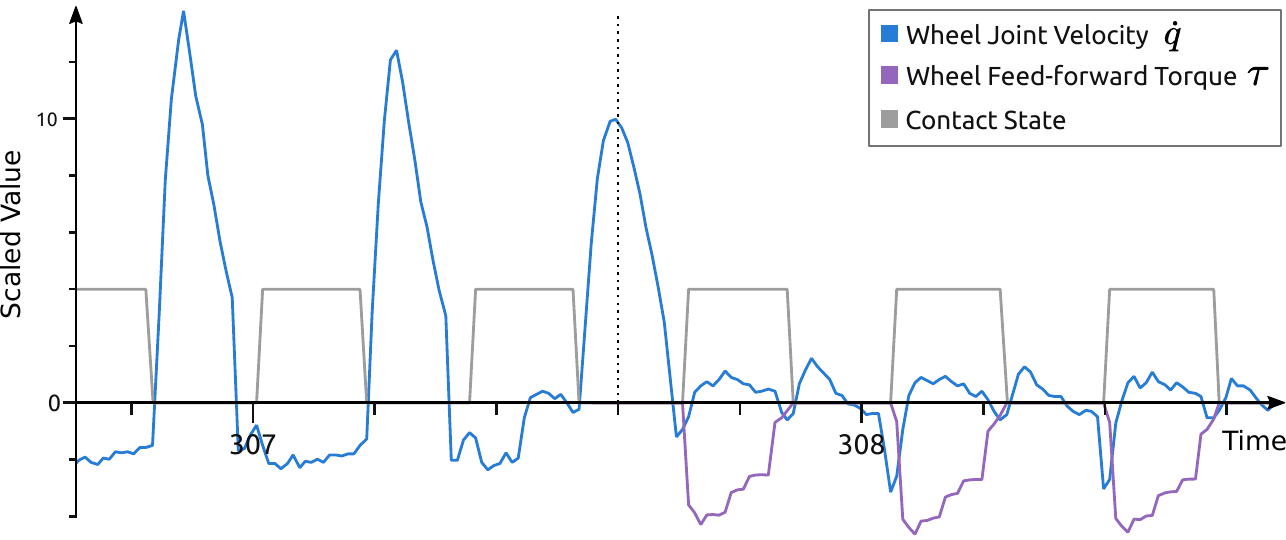}}
	\caption{The feedback effect of the wheel reaction force.
 This graph shows the velocity of the (front-right) wheel joint over time as the robot performs a bounding gait. Initially, the wheel is not receiving any feed-forward torque. After time 307.6 (the vertical dotted line), the wheel receives the torque generated in \myeq{eq_wheel_tau} to counteract the reaction forces applied by the controller, which results in fewer unwanted wheel joint rotations.
 	}
	\label{eval_ss2}
\end{figure}

We evaluated the effect of the wheel controller on the system, shown in \myfig{eval_ss2}. In this test, the robot executes a bounding gait, and the velocity of the front-right wheel is recorded over time. After time \unit[307.6]{s}, the feed-forward torques generated in \myeq{eq_wheel_tau} are applied to the wheel and the wheels work against the reaction force along the $x$ axis to reduce the unnecessary rotations, which results in more stable contact locations along the wheels' rolling direction.

\section{Conclusion}\label{chapter_conclusion}
We incorporated wheels into the kinematic model and state estimator of the quadruped Mini Cheetah robot that we equipped with actively driven wheels, which enabled efficient and versatile hybrid driving-stepping locomotion.
By maintaining a high travel speed with the help of wheels, the robot achieved an increased average locomotion speed.
The effect of the end effector's large geometry was accurately captured to define precise contact points and comprehensive contact velocities.
We formulated a KF for state estimation that includes the robot's position and velocity, as well as the contribution of driving with the wheels into the estimated state, which allowed us to use the control framework of the Mini Cheetah robot with minor modifications.

In the future, we will update the formulation of the MPC and the WBC to accurately perform simultaneous stepping and driving with the wheels.
We will also reduce the robot's transportation cost by forcing the robot to purely drive and perform steps only to respond to disturbances or in response to rough terrain.

\section*{Acknowledgments}
The authors thank Michael Schreiber for fabricating and assembling customized actuated wheels and for his assistance with experimental implementations, and AIS\footnote{Autonomous Intelligent Systems: \url{https://ais.uni-bonn.de}} Lab for the ROS packages used for this project.

\bibliographystyle{IEEEtran}
\bibliography{IEEEabrv,main}

\end{document}